\documentclass[letterpaper, 10 pt, conference]{ieeeconf}  
\IEEEoverridecommandlockouts                              
\usepackage{amsmath,amsfonts}
\usepackage{algorithmic}
\usepackage{algorithm}
\usepackage{array}
\usepackage[caption=false,font=normalsize,labelfont=sf,textfont=sf]{subfig}
\usepackage{textcomp}
\usepackage{stfloats}
\usepackage{url}
\usepackage{verbatim}
\usepackage{graphicx}
\usepackage{cite}
\usepackage{glossaries}
\usepackage{siunitx}
\usepackage{booktabs}
\usepackage{makecell}
\usepackage{multirow}
\newacronym{dvrk}{dVRK}{da Vinci Research Kit}
\newacronym{psm}{PSM}{Patient Side Manipulator}
\newacronym{mtm}{MTM}{Master Tool Manipulator}
\newacronym{ecm}{ECM}{Endoscopic Camera Manipulator}
\newacronym{rcm}{RCM}{Remote Center of Motion}
\newacronym{lnd}{LND}{Large Needle Driver}

\newacronym{ros}{ROS}{Robot Operating System}
\newacronym{urdf}{URDF}{Unified Robot Description Format}
\newacronym{ambf}{AMBF}{Asynchronous Multi-Body Framework}
\newacronym{sam2}{SAM 2}{Segment Anything Model 2}
\newacronym{dlc}{DLC}{DeepLabCut}
\newacronym{vit}{ViT}{Vision Transformer}
\newacronym{mlp}{MLP}{Multilayer Perceptron}

\newacronym{dr}{DR}{Differentiable Rendering}
\newacronym{rl}{RL}{Reinforcement Learning}
\newacronym{cnn}{CNN}{Convolutional Neural Network}
\newacronym{rmis}{RMIS}{Robot-assisted Minimally Invasive Surgery}
\newacronym{dof}{DoF}{Degrees of Freedom}
\newacronym{iou}{IoU}{Intersection over Union}
\newacronym{fov}{FOV}{Field of View}
\newacronym{hsv}{HSV}{Hue, Saturation, and Value}
\newacronym{pf}{PF}{Particle Filter}
\newacronym{dh}{DH parameters}{Denavit-Hartenberg parameters}
\newacronym{pnp}{PnP}{Perspective-n-Point}
\newacronym{mse}{MSE}{Mean Squared Error}
\newacronym{rmse}{RMSE}{Root Mean Squared Error}

\newcommand{\T}[2]{{}^{\mathrm{#1}}\mathbf{T}_{\mathrm{#2}}}

\newcommand{\tildeT}[2]{{}^{\mathrm{#1}}\mathbf{\tilde{T}}_{\mathrm{#2}}}

\begin{document}

\title{\LARGE \bf
Real-time Capable Learning-based Visual Tool Pose Correction via Differentiable Simulation
}

\author{{Shuyuan Yang$^{1}$, Zonghe Chua$^{1}$}%
\thanks{$^{1}$Department of Electrical, Computer, and Systems Engineering, Case Western Reserve University, Cleveland, OH 44106 USA. {\tt\small\{sxy841, zxc703\}@case.edu}}}%



\maketitle

\begin{abstract}
    Autonomy in robot-assisted minimally invasive surgery has the potential to reduce surgeon cognitive and task load, thereby increasing procedural efficiency. However, implementing accurate autonomous control can be difficult due to poor end-effector proprioception. Joint encoder readings are typically inaccurate due to kinematic non-idealities in their cable-driven transmissions. Vision-based pose estimation approaches are highly effective, but lack real-time capability, generalizability, or can be hard to train. 
    
    In this work, we demonstrate a real-time capable, Vision Transformer-based pose estimation approach that is trained using end-to-end differentiable kinematics and rendering. We demonstrate the potential of this approach to correct for noisy pose estimates through a real robot dataset and the potential real-time processing ability. Our approach is able to reduce more than 50\% of hand-eye translation errors in the dataset, reaching the same performance level as an existing optimization-based method. Our approach is four times faster, and capable of near real-time inference at \SI{22}{\hertz}. A zero-shot prediction on an unseen dataset shows good generalization ability, and can be further finetuned for increased performance without human labeling. 
\end{abstract}


\section{Introduction}\label{sec:intro}
\acrfull{rmis} platforms, such as the da Vinci® Surgical System, have seen growing adoption in various \acrshort{rmis} procedures across specialized surgeries~\cite{palep2009robotic}. To reduce the burden on the surgeon, researchers have developed automated approaches for common surgical tasks \cite{saeidi2022autonomous}. 
However, one of the challenges to deploying such autonomous approaches in the real world is the inability to accurately estimate the surgical robot end-effector pose from joint encoders, which results in inaccurate state inputs to autonomous policies or controllers. Such accurate, real-time pose information is also useful in emerging mixed reality \acrshort{rmis} interfaces, in which pose information is visually presented to the user via a digital twin~\cite{9807505,wang2025digital}.

A significant source of the pose error in \acrshort{rmis} systems, such as the \acrfull{dvrk}~\cite{kazanzides2014open}, stems from their cable-driven transmissions. They can produce inaccurate joint angle readings due to friction or cable stretch~\cite{miyasaka2015measurement,haghighipanah2016unscented}, and backlash due to external forces applied to the tool shaft~\cite{cui2023caveats}. This makes computing an accurate hand-eye transform between the end-effector and the stereo endoscope origin for use in planning and control difficult without external sensing approaches. To address this challenge, researchers have developed both model-based~\cite{mahler2014learning} and vision-based~\cite{seita2018fast,richter2021robotic,fan2024reinforcement} methods. 

The above approaches attempt to correct for non-linearities in the forward kinematic chains, thus allowing control in the robot frame. In contrast, the visual servo approach mitigates the uncertainty in the forward kinematics by directly estimating the end-effector pose in the camera frame. This traditionally involves calibrating the hand-eye transform using optical marker tracking~\cite{wu2021closed} of the end-effector, or using an RGBD camera~\cite{hwang2020efficiently,hwang2020applying} for accurate depth perception. However, augmenting the camera, and the end-effectors with markers constrains these methods' applicability to real surgery due to the need for biocompatibility, sterilizability, and miniaturization of wristed instruments. Instead, image-based approaches for tool pose localization have used color-thresholding and neural networks like \acrfull{dlc} \cite{Mathisetal2018} for few-shot finetuned markerless keypoint and feature detection \cite{richter2021robotic,li2020super,lu2021super}. Mask-based approaches leveraging pretrained mask segmentation models such as \acrfull{sam2}~\cite{ravi2024sam2} circumvent the need for manual keypoint labeling, and mitigate issues arising from keypoint ambiguity during tool rotations\,\cite{liang2025differentiable, yang2025instrument}. 

\begin{figure}[!b]
    \centering
    \includegraphics[width=1.0\linewidth]{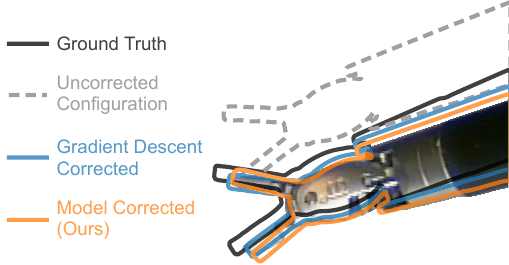}
    \caption{Pose correction of a Large Needle Driver. The colored outlines represent tool pose estimates projected onto the endoscopic image plane. }\label{fig:correct}
\end{figure}

However, unlike with markers, image-based approaches track articulated instruments with partially visible kinematic chains, and attempt to align these features with their current estimates in the camera's image space (see Fig.\,\ref{fig:correct}). Because this alignment depends on inferring projection geometry, the resulting estimation problem becomes non-convex and numerically sensitive, and is typically handled using probabilistic and data-driven vision-based approaches~\cite{richter2021robotic}. The hand-eye transform has been estimated using vision-based \acrfull{pf} approaches \cite{richter2021robotic,hao_vision-based_2018, lu2021super}, and using \acrfull{rl} \cite{fan2024reinforcement}.

By allowing gradients to be backpropagated through the renderer, \acrfull{dr} enables neural networks to be more efficiently trained for 3D and geometric inference tasks~\cite{kato2020differentiable}. \acrshort{dr} has been successfully used for learning-based 3D reconstruction tasks~\cite{yan2016perspective,tulsiani2017multi}. In robotics, Lu et al.~\cite{lu2023image} propose a method that uses \acrshort{dr} to reconstruct the robot end-effector pose by iteratively minimizing the silhouette rendering loss. They further train a network~\cite{lu2023markerless} for robot segmentation and keypoints detection from images via \acrshort{dr}, and then use a PnP solver for articulated robot pose estimation. Yang et al.~\cite{yang2025instrument} use Gaussian Splatting, to reconstruct a 3D surgical tool from a single view and track the tool pose iteratively. Recently, Liang et al.~\cite{liang2025differentiable} develop a \acrshort{dr} approach for \acrshort{dvrk} tool pose calibration using geometric losses without requiring keypoints. 

One significant drawback of existing \acrshort{pf}~\cite{richter2021robotic}, \acrshort{rl}~\cite{fan2024reinforcement}, and \acrshort{dr}~\cite{liang2025differentiable} solutions for the surgical tool pose estimation task is that they are all iterative methods, which require multiple steps to converge on optimal estimates. This feature limits their applicability in real-time control. 

Additionally, there are no open-sourced simulators with end-to-end differentiable kinematics and rendering for the \acrshort{dvrk}, limiting further research into approaches leveraging \acrshort{dr}. For example, `Dr.\ Robot'~\cite{liu2024differentiable} is a new open-sourced \acrshort{dr} robot framework, however, its design is more focused on general articulated robots rather than surgical robotic systems.  


In this letter, we address these identified research gaps through the following contributions: 
\begin{itemize}
  \item A real-time vision-based surgical tool pose estimation neural network model trained using a differentiable visual and kinematic pipeline.
  \item An open-source \acrshort{dvrk}-compatible differentiable kinematics and visual simulation.
  \item A real-world dataset with endoscope camera calibrations and marker-based hand-eye pose labels.
\end{itemize}


\section{Method}\label{sec:method}

\subsection{Differentiable Simulation}\label{subsec:diffsim}
\begin{figure}[!t]
    \centering
    \includegraphics[width=1.0\linewidth]{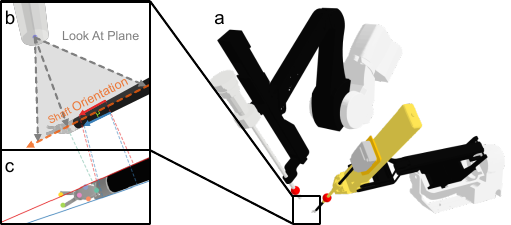}
    \caption{(a) Example scene of the kinematic simulation. (b) Keypoints and the shaft edge lines in the 3D space. (c) Differentiable keypoints and the edge lines projected on the corresponding endoscopic image rendering. }\label{fig:simulator}
\end{figure}

We developed a lightweight kinematic simulator in Python based on PyTorch Kinematics~\cite{Zhong_PyTorch_Kinematics_2024} and PyTorch3D~\cite{ravi2020pytorch3d}. The simulator adopts the same APIs as the SurRoL~\cite{xu2021surrol} \acrshort{dvrk} simulator, and uses the same CAD models~\cite{kazanzides2014open}. The simulator (Fig.~\ref{fig:simulator}a) accepts joint angles and base transforms for both the \acrfull{psm} and the \acrfull{ecm}. A hard Phong shader and a soft silhouette shader~\cite{liu2019soft} are used for differentiable endoscopic image rendering. Relevant computations in the kinematics and rendering pipeline were accelerated with matrix multiplication and broadcasting, and made compatible with batch processing operations. 

As shown in Fig.~\ref{fig:simulator}b, the simulated shaft orientation can be obtained from the kinematics chain. We obtained a direction orthogonal to the shaft axis and the camera image plane, and used it to offset the shaft centerline by its radius, producing two shaft edge lines. We sampled points along these 3D lines and projected them into the image via a fully differentiable camera projection to obtain 2D edge lines. We selected six 3D keypoints (Fig.~\ref{fig:simulator}c) on the end-effector (following prior works~\cite{lu2021super,yang2024vision}), and projected them onto the endoscopic image.

\subsection{Model Architecture}

\begin{figure*}[!ht]
    \centering
    \includegraphics[width=1.0\linewidth]{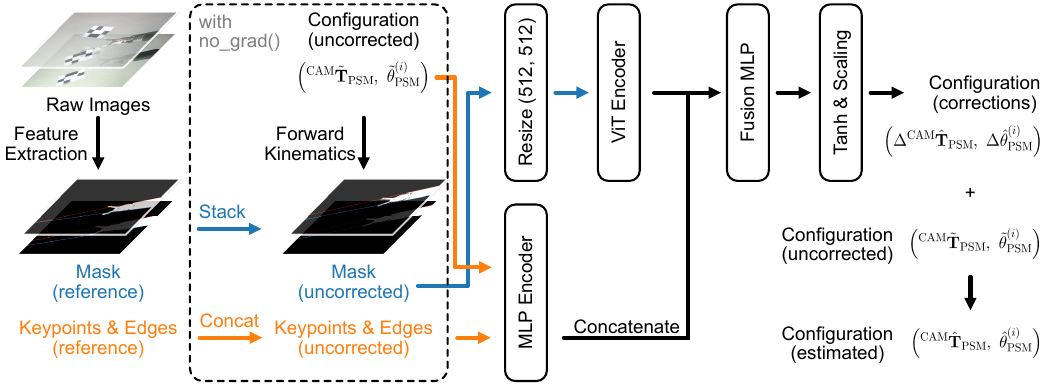}
    \caption{Overview of the proposed vision-based deep learning pose estimation architecture. }\label{fig:arch}
\end{figure*}

We used a \acrfull{vit}~\cite{dosovitskiy2020image} to encode the visual inputs with two pairs of stereo masks being stacked in the channel dimension. As shown in Fig.~\ref{fig:arch}, the visual inputs to the neural network are tool segmentation masks \(\mathbb{M}_{\mathrm{ref}}\) from the stereo endoscopic images, and uncorrected prediction stereo masks \(\tilde{\mathbb{M}}\). In a real scenario, the former corresponds to \acrshort{psm} pose observed under the endoscope, while the latter corresponds to the robot's noisy proprioception, and is derived from the joint encoder and a priori camera-to-robot base transform. This estimated pose is rendered from a projection of the simulated robot.

Due to the partial observability of the kinematic chain, only the last four joints of the \acrshort{psm} (Outer Roll, Wrist Pitch, Wrist Yaw, and End Effector) are visible in the endoscopic view and can be corrected directly. Under this problem formulation, we adopt a similar parametrization as in prior works~\cite{richter2021robotic, yang2025instrument}, lumping the transform from the first observable joint of the \acrshort{psm} chain to the camera origin into a single term. This transform from the camera to \acrshort{psm} Outer Roll frame is parametrized as \( \T{CAM}{PSM} \in \mathrm{SE}(3)\). 

Keypoints \(\mathbf{p}^{(i)}\) and edge lines \(\mathbf{l}^{(i)}\) are extracted from the stereo endoscopic images, and the proprioception-based estimates are projected using \acrshort{dr} in simulation as described above. These numerically represented image features are normalized and concatenated with those of the simulated configuration, then passed through a \acrfull{mlp} encoder. The latent embeddings from the \acrshort{vit} and \acrshort{mlp} encoders are fused using another \acrshort{mlp} with random Dropout and LayerNorm. This layer also enforces the required output dimensionality as the same as the dimension of the parametrized uncorrected configuration \(\in \mathbb{R}^{10}\). A hyperbolic tangent function, multiplied element-wise by learnable scaling factors, is used to bound the outputs within joint limits and the endoscope \acrfull{fov}, thereby stabilizing model training,
\begin{equation}
\Delta \hat{\theta}^{(i)}
= k \,\cdot\, 
\tanh\!\left[
    f\!\left(
        X_{\mathrm{ref}}^{(i)},\ 
        X_{\mathrm{noisy}}^{(i)},\ 
        \tilde{\theta}^{(i)}
    \right)
\right],
\label{eq:correction-2}
\end{equation} 
where \(k\) is a vector of scaling multipliers, and
\begin{equation*}
X_{\mathrm{ref}}^{(i)}=\{\mathbb{M}_{\mathrm{ref}},\mathbf{p}^{(i)},\mathbf{l}^{(i)}\},\
X_{\mathrm{noisy}}^{(i)}=\{\tilde{\mathbb{M}},\tilde{\mathbf{p}}^{(i)},\tilde{\mathbf{l}}^{(i)}\}.
\label{eq:correction-3}
\end{equation*}
The output of (\ref{eq:correction-2}) is used to correct the noisy estimate \(\tilde{\theta}^{(i)}\)
\begin{equation}
    \hat{\theta}^{(i)}
    = \tilde{\theta}^{(i)} + \Delta \hat{\theta}^{(i)}, \label{eq:correction-1} 
\end{equation}
where \(\tilde{\theta}^{(i)} \in \mathbb{R}^{10}\) consists of \(\tildeT{CAM}{PSM}\) comprising a three-dimensional translation vector and three Euler angles; and the four visible joint angles \(\tilde{\theta}^{(i)}_{\mathrm{PSM}}\). 

\subsection{Dataset}
\subsubsection{Data Collection}
A research team member teleoperated both the \acrshort{dvrk} \acrshort{ecm} and a \acrshort{psm} equipped with a Large Needle Driver. Control was alternated between the two to ensure the end-effector was always in view of the endoscope. Teleoperating the \acrshort{ecm} provided poses and viewpoints with realistic and diverse errors for the dataset. 

100 trajectories were recorded at \SI{100}{\hertz} for joint positions and \SI{30}{\hertz} for stereo images in \acrshort{ros} 2 bags.
Joint positions were then downsampled to \SI{30}{\hertz}. The images were recorded using a straight endoscope at \(640 \times 480\) resolution. 

During the post-processing, images were rectified using camera calibration parameters and then passed to \acrshort{dlc} and \acrshort{sam2} for keypoint extraction and tool segmentation, respectively. 100 images were randomly sampled and manually labeled for \acrshort{dlc} training. An initial box prompt was passed to the \acrshort{sam2}.1 large model, as the end-effector always started from the same position for every trajectory. The Hough Line Transform was used on the segmentation mask to detect \acrshort{psm} shaft edge lines, per Richter et al.~\cite{richter2021robotic} 

To quantify the model's error correction performance, a MicronTracker 4 (Claronav, Toronto, ON, Canada) optical tracker was used to measure the pose of the \acrshort{ecm} tip and both of the needle driver jaws. A Kalman filter was used to smooth the optically-tracked poses. 
By averaging the jaws' poses and computing the y-axis components, the ground truth hand-eye transform \(\T{CAM}{EEF}\) and the open jaw angle were measured at a rate of \SI{50}{\hertz}. The noisy observation \(\tilde{\T{CAM}{EEF}}\) was computed using the \acrshort{ecm} and the \acrshort{psm} reported joint angles and a fixed base-to-base transform. 


\subsubsection{Data Split}
\begin{figure}[!t]
    \centering
    \includegraphics[width=1.0\linewidth]{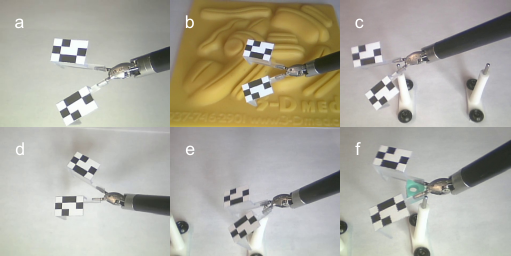}
    \caption{Samples from the dataset. (a)-(c), (e), (f) represent five different scenarios. (a) and (d) are the first frames of the seen and unseen configurations, respectively. }\label{fig:dataset}
    \vspace{-1em}
\end{figure}

The dataset contained five different scenarios: plain background, tissue-like background, pegboard background, pegboard with grasping, and pegboard with ring transfer (Fig.~\ref{fig:dataset}a, b, c, e, and f, respectively). These introduced different scene complexity, such as colored and textured backgrounds, shaft deflections under load, and occlusions. Each scene was used to generate 20 trajectories lasting approximately one minute. A stratified sampling with a \(7:1:2\) ratio was used for the train-validation-test split in each scenario. 

\begin{figure}[!t]
    \centering
    \includegraphics[width=1.0\linewidth]{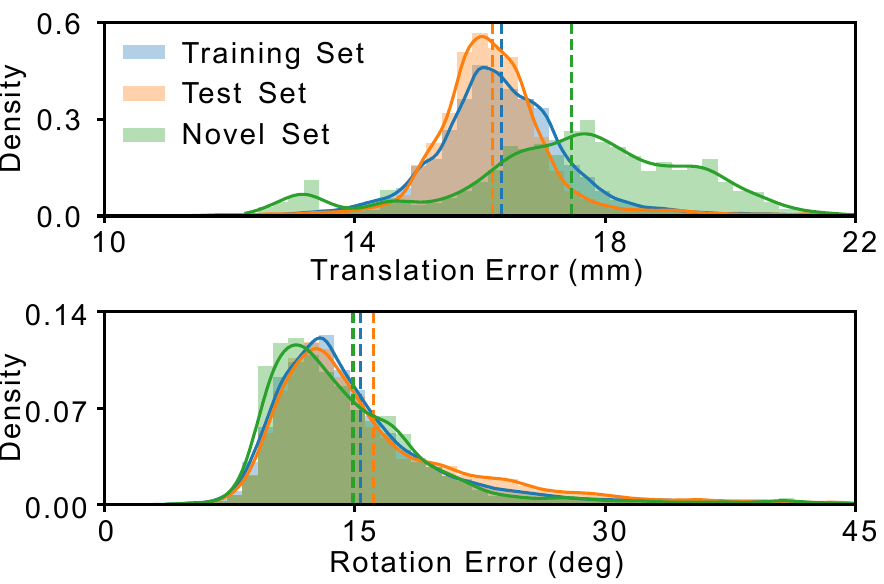}
    \caption{The probability density of normalized hand-eye errors for the training, test, and the novel datasets. }\label{fig:PDF}
    \vspace{-1em}
\end{figure}

To validate the generalization on a novel viewpoint, five extra trajectories were recorded at an unseen configuration. Different base-to-base transforms and different initial joint positions were set for both arms. The stereo cameras were also recalibrated using another batch of images, to introduce a variation in the camera projection matrix. All five unseen trajectories were classified as an extra test set for the generalization experiment in Sec.~\ref{sec:generalization}. The probability density distributions of hand-eye errors (Fig.~\ref{fig:PDF}) across the datasets, highlighting the distributional shift under unseen conditions. 

\subsection{Loss Function}
After each training forward pass, the simulated robot \acrshort{ecm} and \acrshort{psm} is moved to the corrected configuration \(\hat{\theta}^{(i)}\). The simulator then rendered corrected silhouette \(\hat{\mathbb{S}}\), keypoints \(\hat{\mathbf{p}}^{(i)}_{(x,y)}\), and edge lines \(\hat{\mathbf{l}}^{(i)}\) with gradient information for backpropagation. We adopted multiple loss functions to train the model, each computed using different metrics with varying numerical scales. The total loss was
\begin{equation}
    \mathcal{L} = 
    \mathbf{w^T} \cdot 
    \begin{bmatrix}
    \mathcal{L}_{\mathrm{mask}} \\
    \mathcal{L}_{\mathrm{keypoints}} \\
    \mathcal{L}_{\mathrm{shaft}} 
    \end{bmatrix}, 
\label{Eq:loss}
\end{equation}
where \(\mathbf{w}\) is a set of weighted scalars to balance the contributions of each loss term.

\subsubsection{Mask Loss}\label{sec:mask_loss}
The loss is computed as the \acrfull{mse} between the soft silhouette rendered from the corrected configuration and the ground truth mask, plus the absolute difference between their silhouette areas: 
\begin{equation}
    \begin{aligned}
    \mathcal{L}_{\mathrm{mask}} &= 
    \sum_{x=1}^{H} \sum_{y=1}^{W}
    \left(\hat{\mathbb{S}}(x,y) - \mathbb{M}_{\mathrm{ref}}(x,y)\right)^2 \\
    &\quad + \left|
    \sum_{x=1}^{H} \sum_{y=1}^{W} \hat{\mathbb{S}}(x,y)
    - \sum_{x=1}^{H} \sum_{y=1}^{W} \mathbb{M}_{\mathrm{ref}}(x,y)
    \right| ,
    \end{aligned}
    \label{Eq:mask_loss_scale}
\end{equation}
where loss terms reflect the pixel-wise discrepancy between the corrected and true poses, particularly their geometric alignment and scale with respect to the endoscopic view, respectively. While this loss provides informative gradients when the corrected mask overlaps with the ground truth mask, it provides sparse gradients when there is no overlap, which is common during the early stage of the training. 

\subsubsection{Keypoints Loss}\label{sec:keypoint_loss}
The keypoints loss is calculated as the average of Euclidean distances between all corrected and ground truth keypoint pairs:
\begin{equation}
\mathcal{L}_{\mathrm{keypoints}} 
= \frac{1}{N} \sum_{i=1}^{N}
\left\| \hat{\mathbf{p}}^{(i)}_{(x,y)} - \mathbf{p}^{(i)}_{(x,y)} \right\|_2.
\label{Eq:keypoints_loss}
\end{equation}
This loss is useful when the silhouette has no overlap with the mask. Even in the extreme cases where the \acrshort{psm} configuration lies outside the \acrshort{fov}, keypoints can still be represented with negative coordinates for loss computation.

\subsubsection{Shaft Loss}\label{sec:shaft_loss}
We adopt the same polar form representation as Liang et al.~\cite{liang2025differentiable}, where an edge line \(\mathbf{l}^{(i)}\) is divided into \(\{\theta^{(i)}, \rho^{(i)}\}\). This loss computes the average discrepancy of two pairs of edge lines from both angles and distances: 
\begin{equation}
\mathcal{L}_{\mathrm{shaft}} 
= 
\frac{1}{N} \sum_{i=1}^{N}
\left| \hat{\theta}^{(i)} - \theta^{(i)} \right|_{\pi}
+ \gamma \left| \hat{\rho}^{(i)} - \rho^{(i)} \right|,
\label{Eq:line_loss}
\end{equation}
where \(\gamma\) is the weighting factor for the distance term. This loss provides redundant information to the mask loss, but more explicitly enforces shaft alignment. 


\subsection{Learning from Control Priors}
Deploying the model on a new robot with a different error distribution would require finetuning on robot-specific data for maximum performance. To accelerate training and reduce the real-world data burden in these scenarios, we pretrained a model on synthetic data. In simulation, we used a noise injection model~\cite{barragan2024improving} to add error to the \acrshort{psm} joints, while uniform error was introduced to the hand-eye transform. During the pretraining, the scaling vector \(k\) was initially frozen. This ensured that the model provided only small corrections first, enabling the end-effector to stay within view, thus providing more stable gradients for training. Once the backbone encoder outputs stabilized, $k$ was unfrozen, and the model was allowed to learn its optimal value.

\subsection{Environments}
The model was trained on two NVIDIA® L40S GPUs for 100 epochs with early stopping on validation loss. To accelerate training, the training set is randomly downsampled by 80\%. The validation and test sets use the full set without shuffle. Due to the substantial VRAM consumed by \acrshort{dr} pipeline, the batch size was set to 16 and 8 for monocular and stereo models, respectively. 


We use Adam as the optimizer, with both learning rate and weight decay parameters set to 1e-4. The hyperparameters were selected by a grid search performed on a small dataset. Each model was trained three times from different random states, the model with the minimum validation loss was used for inference. Inference was performed on a consumer-grade NVIDIA® RTX 3090 GPU. 

\subsection{Evaluation Metrics}
\subsubsection{\acrlong{rmse}}
We computed the \acrfull{rmse} of corrected poses with respect to the ground truth for each test set. These were averaged over all test sets and reported with a standard deviation. 

\subsubsection{Normalized \acrlong{rmse}}
Besides the \acrshort{rmse}, we compute the N\acrshort{rmse} based on the range of the original data, as 
\begin{equation}
    \mathrm{NRMSE}_i = \frac{\mathrm{RMSE}_i}{y^{(i)}_{\max} - y^{(i)}_{\min}} \times 100\%.
\end{equation}

\subsubsection{Error Reduction Rate}
The Error Reduction Rate is the percent error reduction with respect to the uncorrected observations,
\begin{equation}
    \mathrm{Reduction} = \left( \frac{\mathrm{RMSE}_{\text{uncorr}} - \mathrm{RMSE}_{\text{corr}}}{\mathrm{RMSE}_{\text{uncorr}}} \right) \times 100\%.
\end{equation}


\section{Results and Discussion}\label{sec:exp}

\begin{figure}[b]
    \centering
    \includegraphics[width=1.0\linewidth]{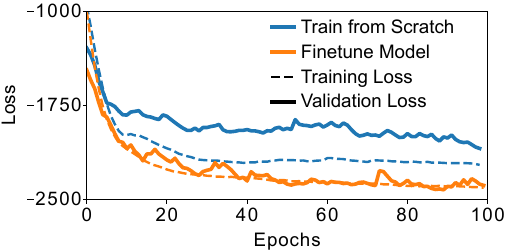}
    \caption{The smoothed training and validation loss plot over the training epochs for two different training approaches: training from scratch and finetuning using pretrained weights. }\label{fig:loss}
\end{figure}

\subsection{Finetuning Approach}
Compared to training from scratch, finetuning using the pretrained weights converged faster and to a lower loss over the first 20 epochs (Fig.~\ref{fig:loss}). The finetuned model also demonstrated less overfitting, and was likely aided by the enhanced data diversity introduced by the  synthetic pretraining dataset. The best validation loss was about 15\% better than training from scratch, within 100 training epochs. 

\subsection{Tool Tracking}
\label{sec:tool_tracking}
\begin{figure*}[t]
    \centering
    \includegraphics[width=1.0\linewidth]{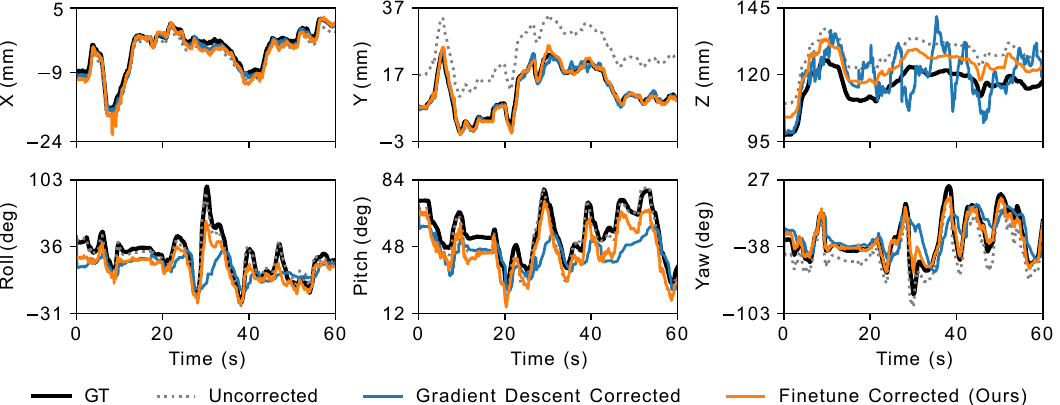}
    \caption{An example of a trajectory from the test set, showing ground truth, uncorrected configuration, and error-corrected trajectories from the gradient descent approach and our finetuned neural network approach corrections, for the end-effector pose.}\label{fig:trajectory}
\end{figure*}

\begin{table}[t]
\centering
\caption{Translation metrics on test sets for monocular scratch and stereo methods.}
\label{tab:translations}
\setlength{\tabcolsep}{5pt}
\begin{tabular}{c c cccc}
\toprule
\textbf{Method} & \textbf{Metric} & \textbf{X} & \textbf{Y} & \textbf{Z} & \textbf{Norm} \\
\midrule

\multirow{6}{*}{\makecell{Scratch \\ (Mono)}}
& \makecell{RMSE \\ (\si{\milli\meter}) $\downarrow$}
& \makecell{3.08 \\ $\pm$0.99}
& \makecell{1.17 \\ $\pm$0.37}
& \makecell{11.56 \\ $\pm$1.54}
& \makecell{11.61 \\ $\pm$1.56} \\[8pt]

& \makecell{NRMSE \\ (\%) $\downarrow$}
& \makecell{11.53 \\ $\pm$4.07}
& \makecell{7.28 \\ $\pm$4.55}
& \makecell{30.70 \\ $\pm$8.40}
& \makecell{30.67 \\ $\pm$8.53} \\[8pt]

& \makecell{Error Reduction \\ (\%) $\uparrow$}
& \makecell{-72.71 \\ $\pm$52.14}
& \makecell{90.18 \\ $\pm$3.12}
& \makecell{-2.26 \\ $\pm$16.85}
& \makecell{9.36 \\ $\pm$14.08} \\
\midrule

\multirow{6}{*}{\makecell{Scratch}}
& \makecell{RMSE \\ (\si{\milli\meter}) $\downarrow$}
& \makecell{2.67 \\ $\pm$1.15}
& \makecell{0.87 \\ $\pm$0.23}
& \makecell{7.21 \\ $\pm$0.69}
& \makecell{7.20 \\ $\pm$0.68} \\[8pt]

& \makecell{NRMSE \\ (\%) $\downarrow$}
& \makecell{10.01 \\ $\pm$4.60}
& \makecell{5.19 \\ $\pm$2.39}
& \makecell{20.51 \\ $\pm$5.95}
& \makecell{20.47 \\ $\pm$6.18} \\[8pt]

& \makecell{Error Reduction \\ (\%) $\uparrow$}
& \makecell{-51.27 \\ $\pm$60.99}
& \makecell{92.71 \\ $\pm$1.96}
& \makecell{36.25 \\ $\pm$7.94}
& \makecell{43.71 \\ $\pm$6.67} \\
\midrule

\multirow{6}{*}{\makecell{Finetune \\ (Ours)}}
& \makecell{RMSE \\ (\si{\milli\meter}) $\downarrow$}
& \makecell{2.49 \\ $\pm$1.15}
& \textbf{\makecell{0.85 \\ $\pm$0.30}}
& \textbf{\makecell{6.32 \\ $\pm$0.71}}
& \textbf{\makecell{6.33 \\ $\pm$0.70}} \\[8pt]

& \makecell{NRMSE \\ (\%) $\downarrow$}
& \makecell{9.30 \\ $\pm$4.62}
& \makecell{4.95 \\ $\pm$2.40}
& \makecell{18.54 \\ $\pm$6.65}
& \makecell{18.49 \\ $\pm$6.84} \\[8pt]

& \makecell{Error Reduction \\ (\%) $\uparrow$}
& \makecell{-39.19 \\ $\pm$57.19}
& \makecell{92.88 \\ $\pm$2.49}
& \makecell{44.36 \\ $\pm$5.81}
& \makecell{50.75 \\ $\pm$5.04} \\
\midrule

\multirow{6}{*}{\makecell{Gradient \\ Descent}}
& \makecell{RMSE \\ (\si{\milli\meter}) $\downarrow$}
& \textbf{\makecell{2.27 \\ $\pm$1.14}}
& \makecell{1.00 \\ $\pm$0.26}
& \makecell{6.68 \\ $\pm$1.47}
& \makecell{6.70 \\ $\pm$1.47} \\[8pt]

& \makecell{NRMSE \\ (\%) $\downarrow$}
& \makecell{8.45 \\ $\pm$4.59}
& \makecell{6.00 \\ $\pm$2.87}
& \makecell{15.70 \\ $\pm$5.94}
& \makecell{15.48 \\ $\pm$5.66} \\[8pt]

& \makecell{Error Reduction \\ (\%) $\uparrow$}
& \makecell{-26.45 \\ $\pm$57.16}
& \makecell{91.65 \\ $\pm$2.19}
& \makecell{41.04 \\ $\pm$14.82}
& \makecell{47.68 \\ $\pm$13.16} \\

\bottomrule
\end{tabular}
\end{table}

\begin{table}[t]
\centering
\caption{Rotation RMSE (\si{\deg}) on test sets for monocular scratch and stereo methods. }
\label{tab:rotations}
\setlength{\tabcolsep}{6pt}

\begin{tabular}{c cccc}
\toprule
\textbf{Method} &
\textbf{Roll} & \textbf{Pitch} & \textbf{Yaw} & \textbf{Norm} \\
\midrule

\makecell{Scratch \\ (Mono)}
& \textbf{\makecell{11.80 \\ $\pm$6.12}}
& \textbf{\makecell{5.63 \\ $\pm$1.04}}
& \textbf{\makecell{9.53 \\ $\pm$5.74}}
& \textbf{\makecell{10.82 \\ $\pm$4.64}} \\[6pt]

\makecell{Scratch}
& \makecell{14.94 \\ $\pm$8.47}
& \makecell{8.80 \\ $\pm$1.27}
& \makecell{10.92 \\ $\pm$8.17}
& \makecell{15.47 \\ $\pm$6.47} \\[6pt]

\makecell{Finetune \\ (Ours)}
& \makecell{15.54 \\ $\pm$8.29}
& \makecell{9.00 \\ $\pm$1.23}
& \makecell{11.51 \\ $\pm$8.02}
& \makecell{16.08 \\ $\pm$6.22} \\[6pt]

\makecell{Gradient \\ Descent}
& \makecell{20.52 \\ $\pm$10.41}
& \makecell{12.63 \\ $\pm$3.93}
& \makecell{18.23 \\ $\pm$9.39}
& \makecell{22.52 \\ $\pm$8.69} \\

\bottomrule
\end{tabular}
\end{table}

Table~\ref{tab:translations} and Table~\ref{tab:rotations} compare the \acrshort{psm} end-effector hand-eye transform \(\T{CAM}{EEF}\) error correction performance of our method, against other variations and a benchmark technique, for translation, and orientation, respectively. 
The norm vectors were calculated across all axes, and then benchmarked with the ground truth for the norm \acrshort{rmse}. 

The monocular (mono) variant of our model was trained from scratch to use the input from the left camera only. Compared to our proposed model that was trained from scratch, it had comparable performance in the x- and y-axes but predictably was unable to substantially resolve the depth (i.e. z-axis) estimate. Compared to training from scratch, the finetuned model showed about 12\% more accuracy on translation. This resulted in an over 50\% error reduction against the uncorrected configuration. 

To benchmark our method, we implemented a gradient descent-based method similar to that described in~\cite{lu2023image,liang2025differentiable}. The method used the same uncorrected configuration and loss functions as our method to iteratively optimize the camera to \acrshort{psm} Roll link transform and the visible four joint angles via Adam. We set a loss convergence threshold equal to our model's average validation loss, and let it converge on each frame rather than initializing a single correction at the start of a task instance as described in~\cite{liang2025differentiable}. The maximum number of iterations allowed for a frame was set to 100, which allowed the optimizer to converge to the reference mask within the first one or two frames and consequently propagate its previous prediction as an initialization for its next iteration. 

Compared to our approach, the gradient descent method showed slightly lower accuracy, particularly in the y- and z-axes, as shown in Table~\ref{tab:translations}.
The differences in error between the two methods were within \SI{1}{\milli\meter}. This sub-millimeter difference is likely below the noise floor of the ground truth of the dataset, leading us to conclude that both methods reached the same level of performance. For the selected test set trajectory shown in Fig.~\ref{fig:trajectory}, both methods produce substantial overlap on the x- and y-axes. However, the gradient descent method had more variability on the z-axis. This was likely due to its error signal oscillating about its convergence threshold.

Notably, all methods showed a negative error reduction on the x-axis, i.e., the pose estimate was less accurate after correction. While large in percentage terms, we note that the base absolute error in that axis was small, at about \SI{3}{\milli\meter}. 

The reduced x-axis accuracy was likely due in part to the inherent calibration bias in the ground truth tracking introduced by the hand-eye calibration performed prior to data collection, which systematically affects all demonstrations within the same dataset. 
Thus, it is possible that in particular scenarios, the marker-based (ground truth) could deviate more from the true pose, than that derived from the joint angle and manual measurements in the x-axis. In Fig.~\ref{fig:correct}, we show that the reprojected ground truth pose can be slightly misaligned with the reference image. This provides evidence to the above, and highlights the difficulty in acquiring accurate ground truth end-effector pose labels for the \acrshort{dvrk}.


According to our results in Table~\ref{tab:rotations}, all methods performed poorly for end-effector orientation correction. Here, the mono model performed the best. This result could have arisen from the discrepancy between the keypoints extracted from the image and the simulator projection. 
Although keypoints projected from the uncorrected pose were used to pre-label the \acrshort{dlc} training data, the final alignment still relied on manual human adjustment, which made it difficult to achieve perfect consistency with the ideal kinematic configuration. Furthermore, the low resolution of the endoscopic images and their limited visual quality introduce additional noise in \acrshort{dlc} inference. Together, these factors lead to discrepancy in the inferred keypoint skeleton, which in turn degrade the performance of rotation correction. 

\subsection{Inference Speed}

\begin{table}[t]
\centering
\caption{Inference performance for mono and stereo methods.}
\label{tab:speed}
\setlength{\tabcolsep}{4pt}

\begin{tabular}{c ccc}
\toprule
\textbf{Mode}
& \textbf{Fwd (\si{\ms}) $\downarrow$} 
& \textbf{Iters $\downarrow$} 
& \textbf{FPS $\uparrow$} \\
\midrule

\multicolumn{4}{c}{\textbf{Ours}} \\
\cmidrule(lr){1-4}

Mono   
& 41.39$\pm$0.84 & \textbf{1} & \textbf{24.17$\pm$0.45} \\[4pt]

Stereo 
& 46.62$\pm$0.57 & \textbf{1} & 21.45$\pm$0.25 \\[6pt]

\multicolumn{4}{c}{\textbf{Gradient Descent}} \\
\cmidrule(lr){1-4}

Mono
& \textbf{22.77$\pm$0.21} & 7.31$\pm$22.69 & 6.01$\pm$18.66 \\[4pt]

Stereo
& 31.67$\pm$0.18 & 11.60$\pm$29.62 & 2.72$\pm$6.95 \\

\bottomrule
\end{tabular}
\end{table}

To evaluate the real-time processing ability of our method, we ran an inference speed experiment on a NVIDIA® RTX 3090 GPU. The inference timing encompassed the entire forward pass. This included the rendering of the uncorrected configuration. The data loading and the correction rendering were excluded from the timing. For comparison, we evaluated the frame rate of the gradient descent-based method. To make this approach run as fast as possible, we applied the same acceleration techniques as our method. The results of these tests are shown in Table~\ref{tab:speed}. 

To simulate a real-time inference scenario, we set the batch size to $1$. The inference speed experiment was performed on 1,800 samples from a trajectory, with an extra 100 samples processed prior to the start of the timer to warm up the GPU. For the gradient descent method, we let it run for up to a maximum of 100 iterations per sample over the same 1,800 samples. The number of iterations reported is the average of across the steps taken per inference on the full test set. 

Although the gradient descent method is slightly faster on a single iteration, the iterative solution requires multiple executions to converge, whereas our method is single-shot deterministic. Hence, compared to the gradient descent method, our method demonstrated approximately $4\times$ faster inference for monocular inputs and $8\times$ faster inference for stereo inputs, which, being close to \SI{30}{\hertz}, is feasible for real-time control at low speeds.

The gradient descent method required a varying number of iterations to meet the alignment threshold. It required 100 iterations at the beginning of the trajectory, and consequently required only a few iterations to track the trajectory. However, the frame rate was highly dependent on the number of iterations required to achieve alignment. Alternatively, Liang et al.~\cite{liang2025differentiable} propose only applying the optimization-based correction once at the beginning of the operation to circumvent the lack of real-time speed. However, in real surgery, there are situations in which these errors do not remain constant. The error distribution approximate a mixture of Gaussians, as shown in Fig.~\ref{fig:PDF}. This could result in drift in one-off calibration approaches. In our particular dataset, we observed that gravity interacted with transmission backlash, exacerbating inaccuracies in the \acrshort{ecm} joint angles as it moved through a steep vertical position. 

\subsection{Ablation Study}

\begin{table}[t]
\centering
\caption{Translation RMSE (\si{\milli\meter}) on test sets under stereo ablations, the baseline is copied from Table~\ref{tab:translations}}
\label{tab:ablation}
\setlength{\tabcolsep}{6pt}
\begin{tabular}{c cccc}
\toprule
\textbf{Ablation} &
\textbf{X} & \textbf{Y} & \textbf{Z} & \textbf{Norm} \\
\midrule

\makecell{Baseline \\ (Scratch)}
    & \makecell{2.67 \\ $\pm$1.15}
    & \makecell{0.87 \\ $\pm$0.23}
    & \makecell{7.21 \\ $\pm$0.69}
    & \makecell{7.20 \\ $\pm$0.68} \\[8pt]

No Masks
    & \makecell{2.49 \\ $\pm$1.22}
    & \makecell{0.92 \\ $\pm$0.30}
    & \makecell{5.36 \\ $\pm$0.66}
    & \makecell{5.37 \\ $\pm$0.65} \\[8pt]

No Keypoints
    & \makecell{32.56 \\ $\pm$6.48}
    & \makecell{116.21 \\ $\pm$1.12}
    & \makecell{72.87 \\ $\pm$2.21}
    & \makecell{113.98 \\ $\pm$3.66} \\[8pt]

No Edges
    & \textbf{\makecell{2.20 \\ $\pm$1.16}}
    & \textbf{\makecell{0.84 \\ $\pm$0.26}}
    & \textbf{\makecell{3.18 \\ $\pm$1.26}}
    & \textbf{\makecell{3.16 \\ $\pm$1.25}} \\

\bottomrule
\end{tabular}
\end{table}

To ensure the features used in our model architecture work as expected, an ablation study was conducted by removing a single feature from the baseline model. Specifically, either the masks, the keypoints, or the edge lines were removed from the inputs, and the corresponding terms loss function were disabled. To adapt the ablation settings, either the \acrshort{vit} encoder was removed or the input dimension of the \acrshort{mlp} encoder was modified. Hence, all ablation models were trained from scratch and benchmarked to the full model, which was also trained from scratch, as in Table~\ref{tab:translations}. 

Table~\ref{tab:ablation} shows the end-effector hand-eye translation \acrshort{rmse}. The model with no keypoints failed at an early stage of training. This was due to the difficulty of constraining the end-effector to remain in the endoscopic \acrshort{fov} without such an explicit loss function. 

Unexpectedly, the model without the edge lines achieved the minimum error, which was less than \SI{4}{\milli\meter} on any axis. Furthermore, the model with no masks and the model with no edge lines displayed lower \acrshort{rmse} than the baseline. As mentioned in Sec.~\ref{sec:mask_loss} and Sec.~\ref{sec:shaft_loss}, both features were able to guide the model towards aligning the tool shaft. However, it is possible that such redundant information is detrimental to the model learning this one-shot control action. For instance, counter-directional gradients from the two terms with respect to the parametrized camera pose may occur during backpropagation, mitigating learning. 

For completeness, we still kept the baseline model containing both features, as the performance degradation was minimal. Redundant information further promotes robustness in cases when a segmentation failure or edge lines are mismatched. In such situations, another feature can serve as a backup for alignment.

\subsection{Generalization to Novel Dataset}
\label{sec:generalization}
\begin{table}[t]
\centering
\caption{Translation RMSE (\si{\milli\meter}) on novel test sets (unseen configuration) using stereo methods.}
\label{tab:novel}
\setlength{\tabcolsep}{6pt}
\begin{tabular}{c cccc}
\toprule
\textbf{Method} & \textbf{X} & \textbf{Y} & \textbf{Z} & \textbf{Norm} \\
\midrule

\makecell{Finetune \\ (Ours)}
& \makecell{2.12 \\ $\pm$1.10}
& \makecell{4.03 \\ $\pm$0.68}
& \textbf{\makecell{2.16 \\ $\pm$0.40}}
& \textbf{\makecell{2.37 \\ $\pm$0.45}} \\[8pt]

\makecell{Gradient \\ Descent}
& \textbf{\makecell{1.37 \\ $\pm$0.14}}
& \textbf{\makecell{2.45 \\ $\pm$0.86}}
& \makecell{10.35 \\ $\pm$1.19}
& \makecell{10.54 \\ $\pm$1.17} \\

\bottomrule
\end{tabular}
\end{table}

To verify the generalization ability of our method, we used the finetuned model for zero-shot inference on a novel dataset collected from an unseen robot configuration. The comparison of translation \acrshort{rmse} with the gradient descent method is shown in Table~\ref{tab:novel}. The correction accuracy on the x- and y-axes was degraded. In contrast, the gradient descent method's accuracy in the x- and y-axes remained stable, as no prior knowledge was required when used on the new dataset. However, our approach showed better generalization in the z-axis compared to the gradient descent method. In fact, the error of our approach on the z-axis was less than that for the seen configuration in Table~\ref{tab:translations}. 
Similar to the negative error reduction in the x-axis discussed in Sec.\,\ref{sec:tool_tracking}, this likely arose from the dataset's intrinsic error distribution.
In this case, the novel dataset could have had a more accurate ground truth pose estimate than that of the seen dataset, which resulted in a lower observed error along the z-axis.

Despite the above limitations, the model showed promising generalization toward the unseen data and error distribution. Moving forward, it is likely that generalization performance can be increased by training with more diverse data across different \acrshort{dvrk} robots. Alternatively, a few-shot finetuning step could be performed to deploy the model on a different robot or to the different \acrshort{ecm} and \acrshort{psm} positions required in different procedures, since the features can be extracted from the image using the proposed pipeline without requiring substantial human labeling. 

\section{Conclusion}\label{sec:con}
In this work, we show a framework that uses the \acrshort{vit} network to estimate the surgical robot end-effector pose using observed image features and uncorrected prediction states. Our experiments demonstrate good performance on the proposed dataset, with good generalization ability to unseen robot configurations. Also, the method displays a promising real-time processing ability over the gold standard iterative differentiable optimization. This allows it to perform pose corrections online rather than through a one-time calibration at the start of the manipulation, making it less susceptible to drift compared to other differentiable approaches.

In future work, we plan to focus on the on-robot applications, using the estimated pose to aid visual control in autonomous telesurgical manipulation. 

\section*{Acknowledgments}
\noindent This work made use of the High Performance Computing Resource in the Core Facility for Advanced Research Computing at Case Western Reserve University. 

\bibliography{bib/lib, bib/references}

@article{ravi2020pytorch3d,
    author = {Nikhila Ravi and Jeremy Reizenstein and David Novotny and Taylor Gordon and Wan-Yen Lo and Justin Johnson and Georgia Gkioxari},
    title = {Accelerating 3D Deep Learning with PyTorch3D},
    journal = {arXiv:2007.08501},
    year = {2020},
}

@software{Zhong_PyTorch_Kinematics_2024,
author = {Zhong, Sheng and Power, Thomas and Gupta, Ashwin and Mitrano, Peter},
doi = {10.5281/zenodo.7700587},
month = feb,
title = {{PyTorch Kinematics}},
version = {v0.7.1},
year = {2024}
}

@article{ravi2024sam2,
  title={SAM 2: Segment Anything in Images and Videos},
  author={Ravi, Nikhila and Gabeur, Valentin and Hu, Yuan-Ting and Hu, Ronghang and Ryali, Chaitanya and Ma, Tengyu and Khedr, Haitham and R{\"a}dle, Roman and Rolland, Chloe and Gustafson, Laura and Mintun, Eric and Pan, Junting and Alwala, Kalyan Vasudev and Carion, Nicolas and Wu, Chao-Yuan and Girshick, Ross and Doll{\'a}r, Piotr and Feichtenhofer, Christoph},
  journal={arXiv preprint arXiv:2408.00714},
  year={2024}
}

@inproceedings{xu2021surrol,
  author={Xu, Jiaqi and Li, Bin and Lu, Bo and Liu, Yun-Hui and Dou, Qi and Heng, Pheng-Ann},
  booktitle={IEEE/RSJ International Conference on Intelligent Robots and Systems}, 
  title={SurRoL: An Open-source Reinforcement Learning Centered and dVRK Compatible Platform for Surgical Robot Learning}, 
  year={2021},
  volume={},
  number={},
  pages={1821-1828},
  doi={10.1109/IROS51168.2021.9635867}
}

@inproceedings{kazanzides2014open,
  author={Kazanzides, Peter and Chen, Zihan and Deguet, Anton and Fischer, Gregory S. and Taylor, Russell H. and DiMaio, Simon P.},
  booktitle={IEEE International Conference on Robotics and Automation}, 
  title={An open-source research kit for the da Vinci® Surgical System}, 
  year={2014},
  volume={},
  number={},
  pages={6434-6439},
  doi={10.1109/ICRA.2014.6907809}
}

@inproceedings{liu2019soft,
  author={Liu, Shichen and Chen, Weikai and Li, Tianye and Li, Hao},
  booktitle={IEEE/CVF International Conference on Computer Vision}, 
  title={Soft Rasterizer: A Differentiable Renderer for Image-Based 3D Reasoning}, 
  year={2019},
  volume={},
  number={},
  pages={7707-7716},
  doi={10.1109/ICCV.2019.00780}
}

@article{Mathisetal2018,
    title = {DeepLabCut: markerless pose estimation of user-defined body parts with deep learning},
    author = {Alexander Mathis and Pranav Mamidanna and Kevin M. Cury and Taiga Abe  and Venkatesh N. Murthy and Mackenzie W. Mathis and Matthias Bethge},
    journal = {Nature Neuroscience},
    year = {2018},
  }

@inproceedings{hao_vision-based_2018,
  author={Hao, Ran and Özgüner, Orhan and Çavuşoğlu, M. Cenk},
  booktitle={IEEE/RSJ International Conference on Intelligent Robots and Systems}, 
  title={Vision-Based Surgical Tool Pose Estimation for the da Vinci® Robotic Surgical System}, 
  year={2018},
  volume={},
  number={},
  pages={1298-1305},
  doi={10.1109/IROS.2018.8594471}
}

@article{kato2020differentiable,
  title={Differentiable rendering: A survey},
  author={Kato, Hiroharu and Beker, Deniz and Morariu, Mihai and Ando, Takahiro and Matsuoka, Toru and Kehl, Wadim and Gaidon, Adrien},
  journal={arXiv preprint arXiv:2006.12057},
  year={2020}
}

@article{liang2025differentiable,
  title={Differentiable Rendering-based Pose Estimation for Surgical Robotic Instruments},
  author={Liang, Zekai and Chiu, Zih-Yun and Richter, Florian and Yip, Michael C},
  journal={arXiv preprint arXiv:2503.05953},
  year={2025}
}

@INPROCEEDINGS{lu2023image,
  author={Lu, Jingpei and Liu, Fei and Girerd, Cédric and Yip, Michael C.},
  booktitle={IEEE International Conference on Robotics and Automation}, 
  title={Image-based Pose Estimation and Shape Reconstruction for Robot Manipulators and Soft, Continuum Robots via Differentiable Rendering}, 
  year={2023},
  volume={},
  number={},
  pages={560-567},
  doi={10.1109/ICRA48891.2023.10161066}
}

@article{yang2025instrument,
  title={Instrument-Splatting: Controllable Photorealistic Reconstruction of Surgical Instruments Using Gaussian Splatting},
  author={Yang, Shuojue and Wu, Zijian and Hong, Mingxuan and Li, Qian and Shen, Daiyun and Salcudean, Septimiu E and Jin, Yueming},
  journal={arXiv preprint arXiv:2503.04082},
  year={2025}
}

@article{dosovitskiy2020image,
  title={An image is worth 16x16 words: Transformers for image recognition at scale},
  author={Dosovitskiy, Alexey and Beyer, Lucas and Kolesnikov, Alexander and Weissenborn, Dirk and Zhai, Xiaohua and Unterthiner, Thomas and Dehghani, Mostafa and Minderer, Matthias and Heigold, Georg and Gelly, Sylvain and others},
  journal={arXiv preprint arXiv:2010.11929},
  year={2020}
}

@article{richter2021robotic,
  title={Robotic tool tracking under partially visible kinematic chain: A unified approach},
  author={Richter, Florian and Lu, Jingpei and Orosco, Ryan K and Yip, Michael C},
  journal={IEEE Transactions on Robotics},
  volume={38},
  number={3},
  pages={1653--1670},
  year={2021},
  publisher={IEEE}
}

@inproceedings{barragan2024improving,
  author={Barragan, Juan Antonio and Ishida, Hisashi and Munawar, Adnan and Kazanzides, Peter},
  booktitle={International Symposium on Medical Robotics}, 
  title={Improving the realism of robotic surgery simulation through injection of learning-based estimated errors}, 
  year={2024},
  volume={},
  number={},
  pages={1-7},
  doi={10.1109/ISMR63436.2024.10585672}
}

@article{cui2023caveats,
  author={Cui, Zejian and Cartucho, João and Giannarou, Stamatia and y Baena, Ferdinando Rodriguez},
  journal={IEEE Robotics \& Automation Magazine}, 
  title={Caveats on the First-Generation da Vinci Research Kit: Latent Technical Constraints and Essential Calibrations [Survey]}, 
  year={2025},
  volume={32},
  number={2},
  pages={113-128},
  doi={10.1109/MRA.2023.3310863}
}

@inproceedings{miyasaka2015measurement,
  author={Miyasaka, Muneaki and Matheson, Joseph and Lewis, Andrew and Hannaford, Blake},
  booktitle={IEEE/RSJ International Conference on Intelligent Robots and Systems}, 
  title={Measurement of the cable-pulley Coulomb and viscous friction for a cable-driven surgical robotic system}, 
  year={2015},
  volume={},
  number={},
  pages={804-810},
  doi={10.1109/IROS.2015.7353464}
}

@inproceedings{haghighipanah2016unscented,
  author={Haghighipanah, Mohammad and Miyasaka, Muneaki and Li, Yangming and Hannaford, Blake},
  booktitle={IEEE International Conference on Robotics and Automation}, 
  title={Unscented Kalman Filter and 3D vision to improve cable driven surgical robot joint angle estimation}, 
  year={2016},
  volume={},
  number={},
  pages={4135-4142},
  doi={10.1109/ICRA.2016.7487606}
}

@article{hwang2020efficiently,
  author={Hwang, Minho and Thananjeyan, Brijen and Paradis, Samuel and Seita, Daniel and Ichnowski, Jeffrey and Fer, Danyal and Low, Thomas and Goldberg, Ken},
  journal={IEEE Robotics and Automation Letters}, 
  title={Efficiently Calibrating Cable-Driven Surgical Robots With RGBD Fiducial Sensing and Recurrent Neural Networks}, 
  year={2020},
  volume={5},
  number={4},
  pages={5937-5944},
  doi={10.1109/LRA.2020.3010746}
}

@inproceedings{hwang2020applying,
  author={Hwang, Minho and Seita, Daniel and Thananjeyan, Brijen and Ichnowski, Jeffrey and Paradis, Samuel and Fer, Danyal and Low, Thomas and Goldberg, Ken},
  booktitle={International Symposium on Medical Robotics}, 
  title={Applying Depth-Sensing to Automated Surgical Manipulation with a da Vinci Robot}, 
  year={2020},
  volume={},
  number={},
  pages={22-29},
  doi={10.1109/ISMR48331.2020.9312948}
}

@article{wu2021closed,
  author={Wu, Baibo and Wang, Longfei and Liu, Xu and Wang, Linhui and Xu, Kai},
  journal={IEEE Robotics and Automation Letters}, 
  title={Closed-Loop Pose Control and Automated Suturing of Continuum Surgical Manipulators With Customized Wrist Markers Under Stereo Vision}, 
  year={2021},
  volume={6},
  number={4},
  pages={7137-7144},
  doi={10.1109/LRA.2021.3097260}
}

@article{li2020super,
  author={Li, Yang and Richter, Florian and Lu, Jingpei and Funk, Emily K. and Orosco, Ryan K. and Zhu, Jianke and Yip, Michael C.},
  journal={IEEE Robotics and Automation Letters}, 
  title={SuPer: A Surgical Perception Framework for Endoscopic Tissue Manipulation With Surgical Robotics}, 
  year={2020},
  volume={5},
  number={2},
  pages={2294-2301},
  doi={10.1109/LRA.2020.2970659}
}

@inproceedings{lu2021super,
  author={Lu, Jingpei and Jayakumari, Ambareesh and Richter, Florian and Li, Yang and Yip, Michael C.},
  booktitle={IEEE International Conference on Robotics and Automation}, 
  title={SuPer Deep: A Surgical Perception Framework for Robotic Tissue Manipulation using Deep Learning for Feature Extraction}, 
  year={2021},
  volume={},
  number={},
  pages={4783-4789},
  doi={10.1109/ICRA48506.2021.9561249}
}

@article{fan2024reinforcement,
  author={Fan, Ke and Chen, Ziyang and Liu, Qiaoling and Ferrigno, Giancarlo and Momi, Elena De},
  journal={IEEE Transactions on Medical Robotics and Bionics}, 
  title={A Reinforcement Learning Approach for Real-Time Articulated Surgical Instrument 3-D Pose Reconstruction}, 
  year={2024},
  volume={6},
  number={4},
  pages={1458-1467},
  doi={10.1109/TMRB.2024.3464089}
}

@article{yan2016perspective,
  title={Perspective transformer nets: Learning single-view 3d object reconstruction without 3d supervision},
  author={Yan, Xinchen and Yang, Jimei and Yumer, Ersin and Guo, Yijie and Lee, Honglak},
  journal={Advances in Neural Information Processing Systems},
  volume={29},
  year={2016}
}

@inproceedings{tulsiani2017multi,
  author={Tulsiani, Shubham and Zhou, Tinghui and Efros, Alexei A. and Malik, Jitendra},
  booktitle={IEEE Conference on Computer Vision and Pattern Recognition}, 
  title={Multi-view Supervision for Single-View Reconstruction via Differentiable Ray Consistency}, 
  year={2017},
  volume={},
  number={},
  pages={209-217},
  doi={10.1109/CVPR.2017.30}
}

@inproceedings{lu2023markerless,
  title={Markerless Camera-to-Robot Pose Estimation via Self-Supervised Sim-to-Real Transfer},
  author={Lu, Jingpei and Richter, Florian and Yip, Michael C},
  booktitle={IEEE/CVF Conference on Computer Vision and Pattern Recognition},
  pages={21296--21306},
  year={2023}
}

@article{liu2024differentiable,
  title={Differentiable robot rendering},
  author={Liu, Ruoshi and Canberk, Alper and Song, Shuran and Vondrick, Carl},
  journal={arXiv preprint arXiv:2410.13851},
  year={2024}
}

@inproceedings{seita2018fast,
  author={Seita, Daniel and Krishnan, Sanjay and Fox, Roy and McKinley, Stephen and Canny, John and Goldberg, Ken},
  booktitle={IEEE International Conference on Robotics and Automation}, 
  title={Fast and Reliable Autonomous Surgical Debridement with Cable-Driven Robots Using a Two-Phase Calibration Procedure}, 
  year={2018},
  volume={},
  number={},
  pages={6651-6658},
  doi={10.1109/ICRA.2018.8460583}
}

@inproceedings{mahler2014learning,
  author={Mahler, Jeffrey and Krishnan, Sanjay and Laskey, Michael and Sen, Siddarth and Murali, Adithyavairavan and Kehoe, Ben and Patil, Sachin and Wang, Jiannan and Franklin, Mike and Abbeel, Pieter and Goldberg, Ken},
  booktitle={IEEE International Conference on Automation Science and Engineering}, 
  title={Learning accurate kinematic control of cable-driven surgical robots using data cleaning and Gaussian Process Regression}, 
  year={2014},
  volume={},
  number={},
  pages={532-539},
  doi={10.1109/CoASE.2014.6899377}
}

@article{palep2009robotic,
  title={Robotic assisted minimally invasive surgery},
  author={Palep, Jaydeep H},
  journal={Journal of Minimal Access Surgery},
  volume={5},
  number={1},
  pages={1--7},
  year={2009},
  publisher={Medknow}
}

@article{yang2024vision,
author = {Yang, Shuyuan and Le, My H. and Golobish, Kyle R. and Beaver, Juan C. and Chua, Zonghe},
title = {Vision-Based Force Estimation for Minimally Invasive Telesurgery Through Contact Detection and Local Stiffness Models},
journal = {Journal of Medical Robotics Research},
volume = {09},
number = {03n04},
pages = {2440008},
year = {2024},
doi = {10.1142/S2424905X24400087},
}

@INPROCEEDINGS{9807505,
  author={Long, Yonghao and Cao, Jianfeng and Deguet, Anton and Taylor, Russell H. and Dou, Qi},
  booktitle={International Symposium on Medical Robotics}, 
  title={Integrating Artificial Intelligence and Augmented Reality in Robotic Surgery: An Initial dVRK Study Using a Surgical Education Scenario}, 
  year={2022},
  volume={},
  number={},
  pages={1-8},
  doi={10.1109/ISMR48347.2022.9807505}
}

@INPROCEEDINGS{wang2025digital,
  author={Wang, Junxiang and Barragan, Juan Antonio and Ishida, Hisashi and Guo, Jingkai and Ku, Yu-Chun and Kazanzides, Peter},
  booktitle={International Symposium on Medical Robotics}, 
  title={A Digital Twin for Telesurgery Under Intermittent Communication}, 
  year={2025},
  volume={},
  number={},
  pages={218-224},
  doi={10.1109/ISMR67322.2025.11025988}
}

@article{saeidi2022autonomous,
  title={Autonomous robotic laparoscopic surgery for intestinal anastomosis},
  author={Saeidi, Hamed and Opfermann, Justin D and Kam, Michael and Wei, Shuwen and L{\'e}onard, Simon and Hsieh, Michael H and Kang, Jin U and Krieger, Axel},
  journal={Science Robotics},
  volume={7},
  number={62},
  pages={eabj2908},
  year={2022},
  publisher={American Association for the Advancement of Science}
}
\bibliographystyle{IEEEtran}

\end{document}